\def\year{2019}\relax
\documentclass[letterpaper]{article} 
\usepackage{aaai19} 
\usepackage{times} 
\usepackage{helvet} 
\usepackage{courier} 
\usepackage{url} 
\usepackage{graphicx} 

\usepackage{subfigure}

\usepackage{lipsum}

\begin{document}
%
\title{Understanding the Importance of Single Directions \\ via Representative Substitution}

\author{Li Chen$^\dagger$, Hailun Ding$^\ddagger$, Qi Li$^\uparrow$, Zhuo Li$^\uparrow$, Jian Peng$^\dagger$, Haifeng Li$^{\dagger*}$ \\ $^\dagger$School of Geosciences and Info-Physics , $^\ddagger$School of Software\\$^\uparrow$School of Information Science and Engineering \\ Central South University \\  Changsha Hunan 410083 P.R. China \\ \{vchenli, hlding, dsjliqi, pengj2017, lihaifeng\}@csu.edu.cn,buebirzhuo@gmail \\ $^*$Corresponding author}

\maketitle
\begin{abstract}
Understanding the internal representations of deep neural networks (DNNs) is crucal to explain their behavior. The interpretation of individual units, which are neurons in MLPs or convolution kernels in convolutional networks, has been paid much attention given their fundamental role. However, recent research \cite{morcos2018importance} presented a counterintuitive phenomenon, which suggests that an individual unit with high class selectivity, called interpretable units, has poor contributions to generalization of DNNs. In this work, we provide a new perspective to understand this counterintuitive phenomenon, which makes sense when we introduce \textit{Representative Substitution} (RS). Instead of individually selective units with classes, the RS refers to the independence of a unit's representations in the same layer without any annotation. Our experiments demonstrate that interpretable units have high RS which are not critical to network's generalization. The RS provides new insights into the interpretation of DNNs and suggests that we need to focus on the independence and relationship of the representations.

\end{abstract}

\section{Introduction}
Deep neural networks (DNNs) have made remarkable progress in various domains, such as image classification, word recognition, and language translation. Despite the substantial success of DNNs, their working mechanism, such as the capability to memorize the entire datasets with low generalization error, is still an open issue.

Inspired by neuroscience, recent works \cite{morcos2018importance} have demonstrated \textit{high interpretability units} (HIUs), which are important response to a class of objects and were believed to play a crucial role in the performance of DNNs in the past, actually have an inconsiderable contribution to accuracy and even damage a network's generalization performance. Instead, \textit{weak interpretability units} (WIUs) may be important to improve the accuracy and generalization of DNNs. Different experiments, such as ablation and perturbation, show that a single direction has minimal importance to generalization. Therefore, previous studies of understanding the working mechanism of a network by analyzing HIUs may be misleading \cite{radford2017learning,nguyen2016synthesizing,zeiler2014visualizing}. The relationship among HIUs, WIUs, accuracy, and generalization remains unclear.

Regardless of whether a unit is sensitive to class, latent patterns are found with the corresponding unit when the unit activation is large. Therefore, we take the activation maximization (AM) \cite{visualization_techreport} of units as their representation in the feature space. This representation propagates the image initialized with random noise to compute the activation of certain unit and then passes the activation backward to update the best direction of the input image. This method prevents the impacts of class and limited data samples in the activation space. We propagate the generated image forward the network and record all activation in the same layer. Although the generated image is optimized for the target unit, many units' activation is larger than that of a certain unit. In this optimization direction, other units also have the same representations. 

We define \textit{Representative Substitution} (RS), the ratio of units with activation greater than that of a target unit to the full layer, to quantify the independence of a unit's representations in the same layer for understanding the behavior of HIUs and WIUs and the influence of their interactions on the accuracy and generalization of DNNs. We also improve AM to obtain independent representations, which reduces the influences of other units in the image generation process. Regardless of what generates an image, the HIUs' RS is relatively high. When a unit is ablated, a large number of other units can complement the loss representations in the same layer; therefore, HIUs are inconsiderably important. We find why the relationship between class selectivity and unit importance is negatively correlated or irrelevant. The first few layers of the network are for feature extraction, whereas the remaining layers are for feature integration. HIUs and WIUs' importance differs in different layers. The main contributions are summarized as follows:

(1) We forge a concept, called \textit{Representative Substitution}, to indicate the independence of a unit's representations in the same layer. The importance of HIUs to the network is explained through the perspective of feature space.

(2) The importance of HIUs is not a universal phenomenon; instead, it heavily depends on the layers they are located given different convolutional layers have different functions, feature extraction, and feature integration. 

(3) Our method is built on the representations of convolution kernels without any information on classes or labels, which give us a potential opportunity to analyze the intrinsic features of convolution kernels. Our research suggests that independent representations of convolution kernels may be a promising way to explain the behavior of DNNs.

\section{Methods} 
We propose an \textit{independent activation maximization} (IAM) which represents a unit in the feature space, and \textit{Representative Substitution} (RS) which represents the independence of the unit's representations in the same layer, to understand the characteristics of HIUs and WIUs and their impacts on the network importance.

We introduce basic concepts before the IAM and RS are discussed.

\noindent\textbf{Class Selectivity.} A metric is inspired by system neuroscience to quantify the class selectivity of units \cite{morcos2018importance}. The selectivity index is calculated as follows:

\begin{equation}\label{sel}
    selectivity(i) = \frac{\bar{x}^i_{max}-\bar{x}^i_{-max}}{\bar{x}^i_{max}+\bar{x}^i_{-max}},
\end{equation}

where $\bar{x}^i_{max}$ is the maximum mean activation of the $i$-th unit under class and $\bar{x}^i_{-max}$ is the mean activation of the $i$-th unit across all other classes. The $i$-th unit mean activation of all samples under each class is calculated, and the vector of dimension equals to the number of classes. The class selectivity of the $i$-th unit can be obtained by Eq.\ref{sel}. The metric varies from 0 to 1; 0 denotes that a unit activates identically for all classes, whereas 1 denotes that a unit only activates for a single class. In other words, a unit with high class selectivity value is an HIU and a unit with low class selectivity value is a WIU. 

\smallskip
\noindent\textbf{Activation Maximization.} As shown in \cite{visualization_techreport}, the purpose of the algorithm is to visualize the representations of a network's unit after the end of training. The \textit{activation maximization} (AM) of the $i$-th unit is  

\begin{equation}
    AM(i) = \arg\max {x^i}, 
\end{equation}

where $x^i$ is the activation of the $i$-th unit. In detail, we put a random noise image forward the network to compute the activation of the $i$-th unit and then propagate activation backward through the network to compute the gradient of the input image. After a few iterations, we can acquire the generated image; thus, the activation of the $i$-th unit becomes large. The generated image is considered as visual representations of the unit in the feature space.

\smallskip
\noindent\textbf{Independent Activation Maximization.} In the process of image generation, the AM algorithm only follows consideration of the maximum activation value of the target unit. However, other units may also have high activation to the input image, which makes the visual representations entangled with other units. Thus, we modify the objective function to generate the following IAM formula,

\begin{equation}\label{iam}
    IAM(i) = \arg\max {(x^i - \bar x^{-i})}, 
\end{equation}

where $\bar x^{-i}$ is the mean activation of all units without the $i$-th unit in the same layer. The images are obtained by Eq.\ref{iam}, which can maximize the activation value of the $i$-th unit while the activation of other units is suppressed. Therefore, images become the independent visual representations of the unit.

\smallskip
\noindent\textbf{Representative Substitution.} We generate an image for a certain unit by the IAM algorithm to obtain the unit’s representations and then pass the generated image to the network for quantifying the impacts of other units on the representation of the target unit in the same layer. The activation of all units in the same layer of a certain unit is recorded. The RS is defined by the number of units greater than the activation of the certain unit divided by the total number of units in that layer. The formula is

\begin{equation}
    RS(i) = \frac{|\{x^{'}:x^{'} > x^{i} \} |}{|\{x \} |} \quad s.t. \quad x^{i} = f(IAM(i)), 
\end{equation}

where $|\{x \} |$ represents the number of activation or units and $f(image)$ represents networks' feedforward computation that the image can be obtained by IAM. The RS indicates the degree to which visual representations of a unit is activated by other units in the same layer. This metric varies from 0 to 1; 0 indicates that the visual representations of the $i$-th unit is independent and unique, whereas 1 indicating the all units have the same representations with the $i$-th unit.

\section{Experiments}
\noindent\textbf{Experimental settings} 

In the experiments, we use various DNNs, including an MLP, a shallow convolutional network, VGG16 \cite{Simonyan2014Very} under CIFAR-10 \cite{Krizhevsky2009Learning}, and ImageNet \cite{Deng2009ImageNet} datasets. Each layer of MLP contains 128, 512, 2048 and 2048 neurons. The shallow convolutional network is trained on CIFAR-10 for 100 epochs. Its convolutional layer sizes are 64, 64, 128, and 128. All kernels are 3$\times$3, with strides of 1. CIFAR-10 contains 10 classes with 60,000 images, and ImageNet has 1.2 million images from 1,000 classes. 

\smallskip
\noindent\textbf{Relationship with class selectivity}

\begin{figure}[tbp]
  \centering
  \subfigure[layer1]{
    \label{fig:a1} 
    \includegraphics[scale=0.22]{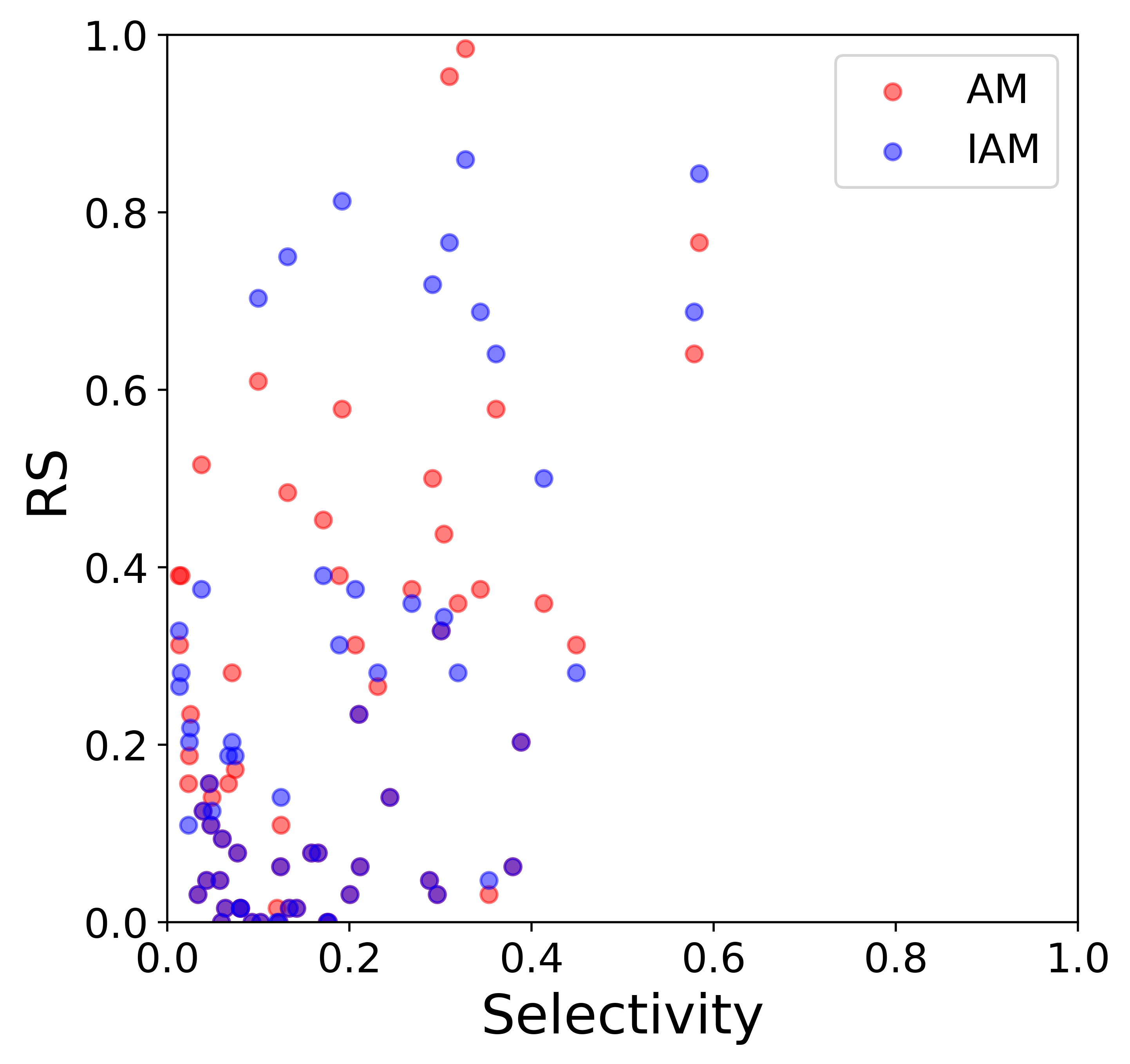}}
  \subfigure[layer3]{
    \label{fig:a3}
    \includegraphics[scale=0.22]{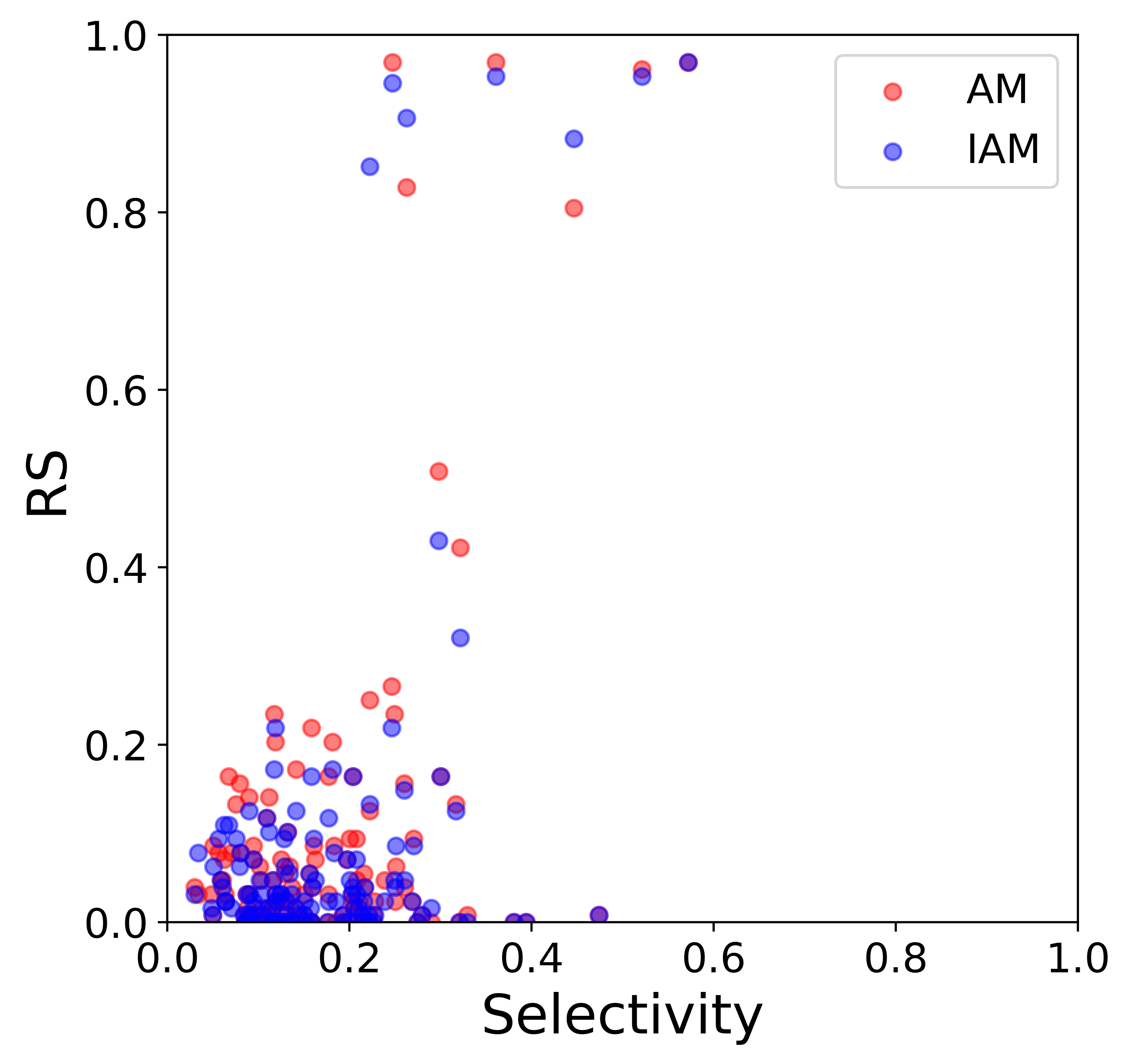}}
  \caption{\textbf{The RS value of HIUs is high, whereas that RS value of WIUs is low. Each point represents a convolution kernel.} Relation between RS and class selectivity under AM and IAM algorithms in the first layer of the shallow convolutional network (a) and the third layer (b).}
  \label{fig:vgg_1_3} 
\end{figure}

We train a shallow convolutional network on Cifar-10 and compute the class selectivity of all units. We then obtain the RS results under the IAM and the AM algorithms for comparison, and select two layers of the network, as shown in Fig.\ref{fig:vgg_1_3}. The experimental results show that the RS is high when the class selectivity of the convolution kernels in the two layers is high.

The WIUs with low class selectivity also have low RS. A low RS indicates independent and irreplaceable representations in the same layer. Therefore, the WIUs are important to network’s generalization. When the RS is obtained by the AM algorithm, the lower left corner of the figure is dense, whereas the upper right corner is sparse. This finding shows that the representations of WIUs in this layer are difficult to replace, whereas HIUs are easy to replace. We use the improved AM algorithm, the IAM algorithm, to compute the units’ RS. Although the overall RS declines slightly, the basic trend is consistent with that of AM, and the same conclusions are reached. HIUs are unimportant because other alternative representations exist. WIUs represent several independent features, and its ablation largely impacts the network generalization.

\begin{figure}[tbp]
  \centering

   \includegraphics[scale=0.66]{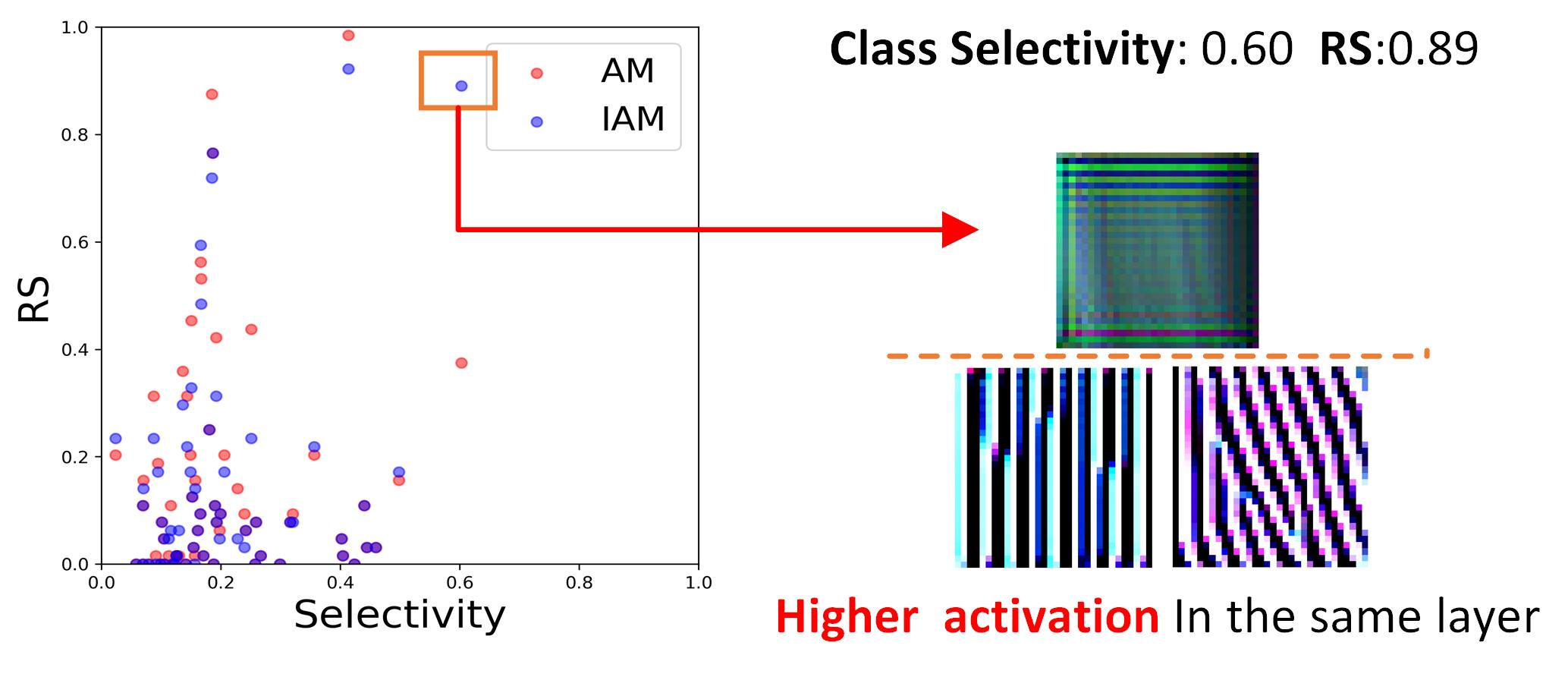}
 
  \caption{\textbf{The HIU's representations are simple compared with the representations of other units.} The HIU has high class selectivity and RS.}
  \label{fig:vis} 
\end{figure}

We visualize an HIU and two units with higher activation than the HIU by the IAM algorithm, as shown in Fig.\ref{fig:vis}. The representations of other units cover the HIU's representations, but they are also given their own independent features. Thus, HIUs with simple representations are unsuitable for network generalization. In the latter experiment, we utilize the IAM algorithm to acquire the RS.

\bigskip
\noindent\textbf{Relationship with different layers}
\smallskip

\begin{figure}[tbp]
  \centering
  \subfigure[fully connected layer before the classifier]{
    \label{fig:vggful} 
    \includegraphics[scale=0.22]{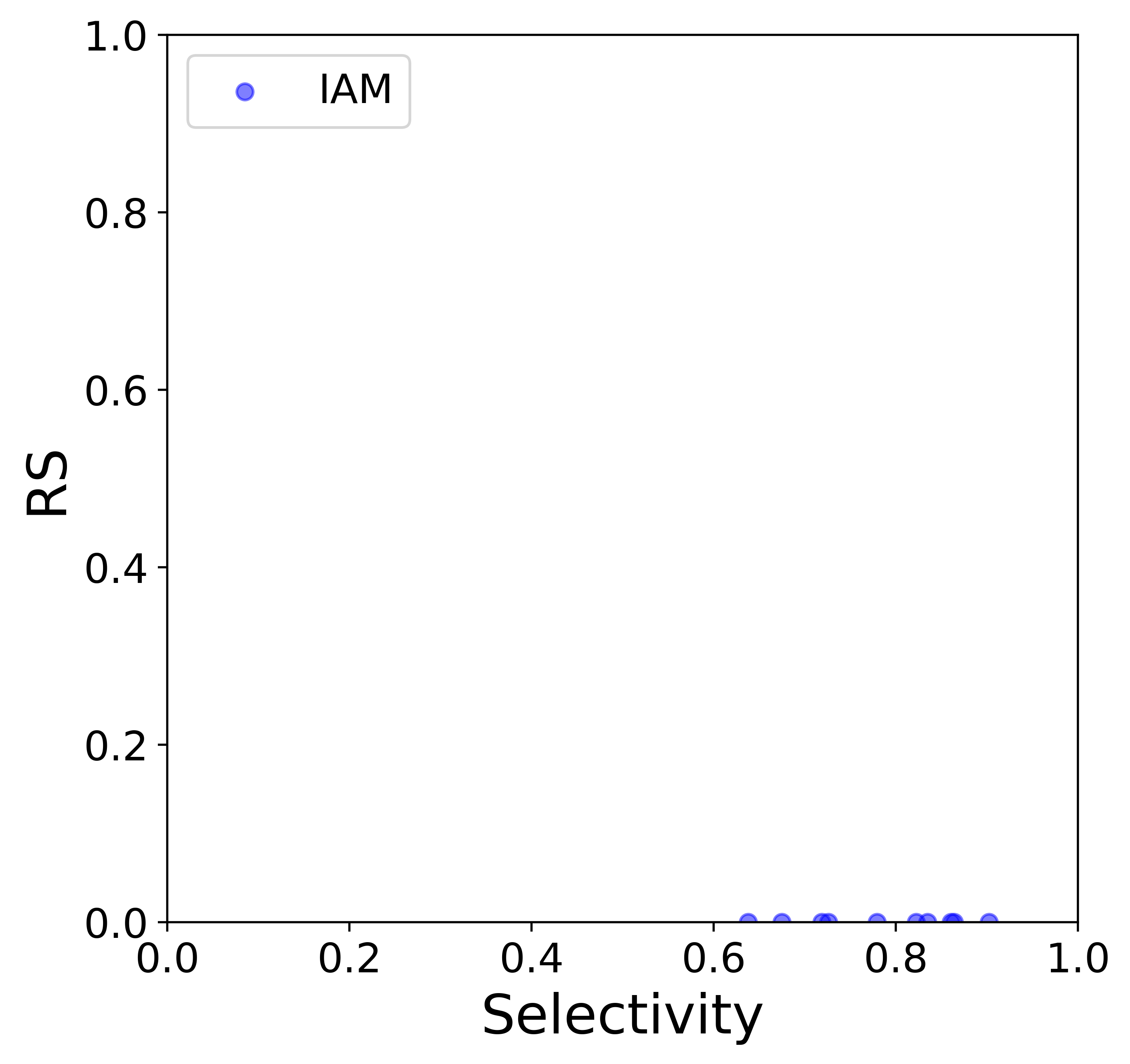}}
  \subfigure[correlation between RS and class selectivity under different layers]{
    \label{fig:tend} 
    \includegraphics[scale=0.22]{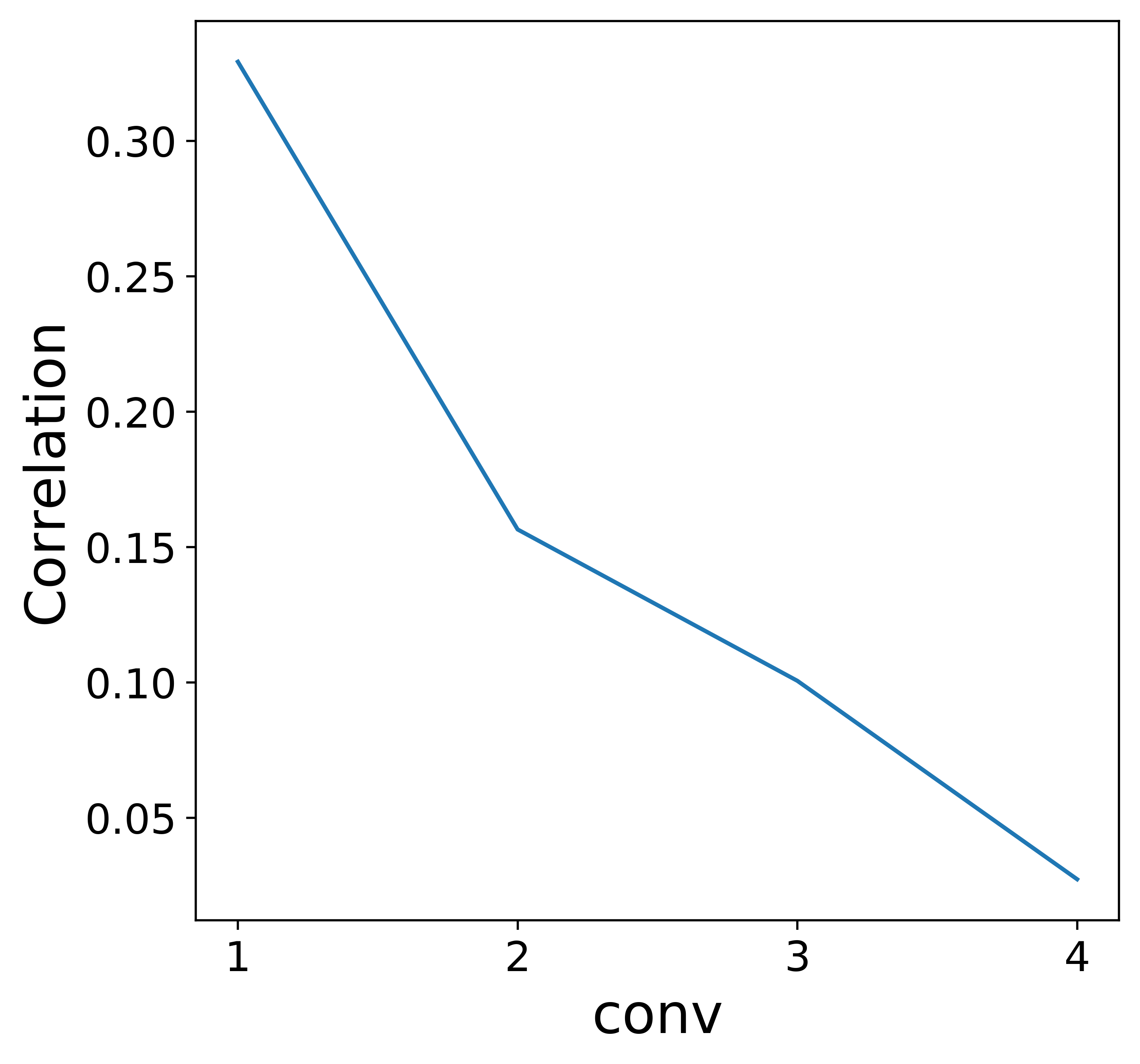}}
  \label{fig:vgg4_c_t} 
  \caption{\textbf{ The class selectivity and RS have different correlations on different layers.} (a) The units' class selectivity is high and the RS is low in the fully connected layer. (b) The correlation between class selectivity and RS decreases as the number of layers increases.}
\end{figure}

Unfortunately, a unit of a good network must be a single class on the fully connected layer before the classifier. Although a HIU is given, its representations may not be substituted in the same layer. Therefore, we compute the RS and class selectivity of the fully connected layer of the shallow convolutional network, as shown in Fig.\ref{fig:vggful}. The results indicate that the class selectivity of units in the fully connected layer before the classifier is large, and the RS is mostly zero.

HIUs are consequently inappropriate to indicate the quality of network generalization. The importance of HIUs is related to which layer it locates. We conduct a Spearman’s correlation analysis of the RS and class selectivity of each convolutional layer of the shallow convolutional network, as shown in Fig.\ref{fig:tend}. The results demonstrate that the RS and class selectivity are positively correlated, and the correlation gradually decreases as the number of layers increases. So in \cite{morcos2018importance}, the importance of unit ablation under different convolutional layers on the network is thus difficult to determine because the way of ablation is random. From a feature space prospective, as the number of layers increases, the independence of features on the unit gradually becomes unrelated to the interpretability of the unit. \citeauthor{morcos2018importance} clarified that class selectivity has a strong relationship with importance in the shallow layer, but it gradually decays as layer deepens. The first few layers of the convolutional network have feature extraction capabilities; hence, the HIUs in these layers are unimportant because their RS are high and other convolution kernels can replace their representations. Therefore, too many HIUs will reduce the generalization performance of the network. As the number of layers increases, feature extraction is no longer part of the main influencing factor of network performance. The network needs to integrate features and provide semantic information. This finding is consistent with theoretical observations \cite{raghu2016expressive}.

\begin{figure}[tbp]
  \centering
  \subfigure[shallow layers]{
    \label{fig:subfig:a} 
    \includegraphics[scale=0.22]{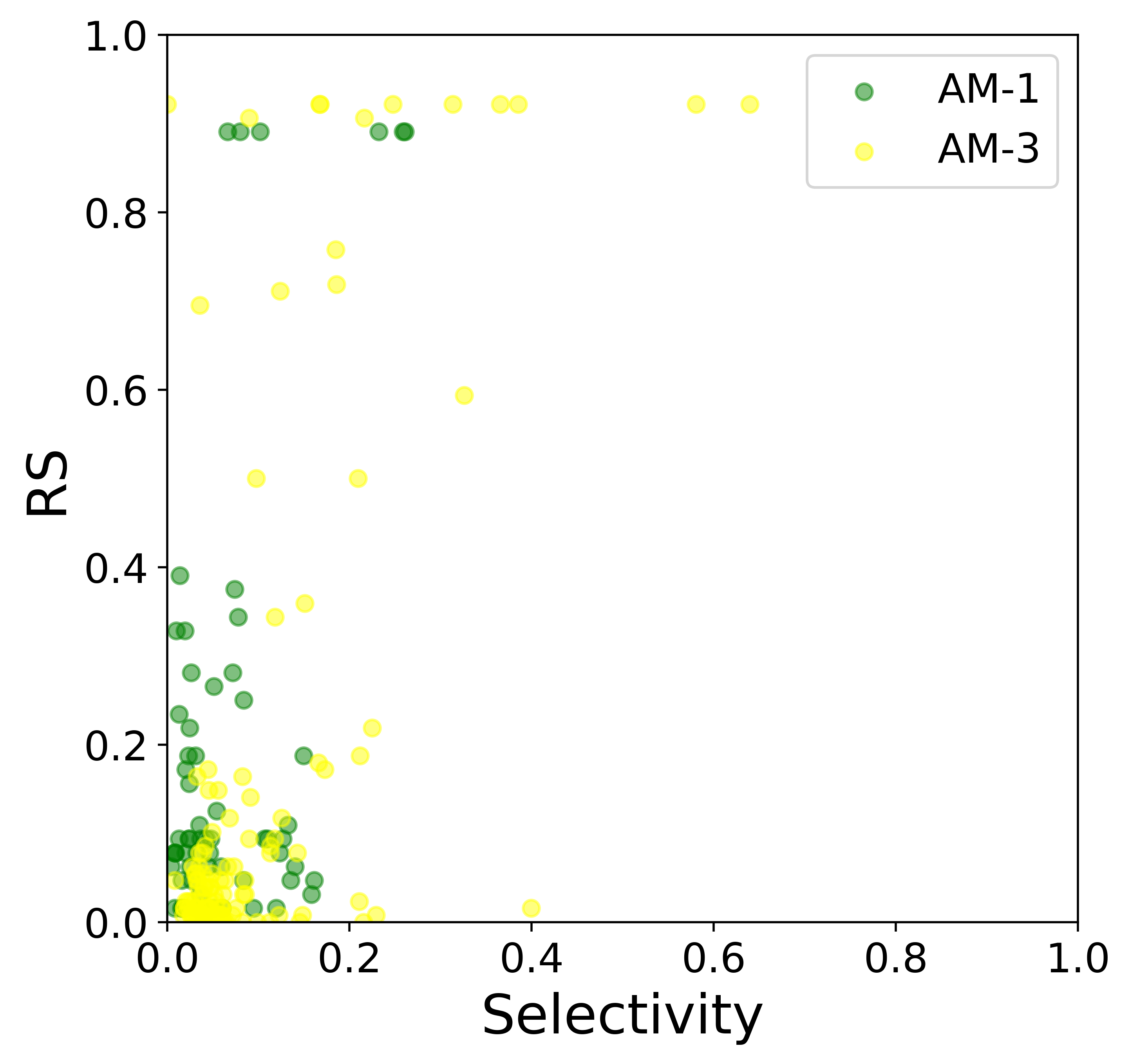}}
  \subfigure[deep layers]{
    \label{fig:subfig:b} 
    \includegraphics[scale=0.22]{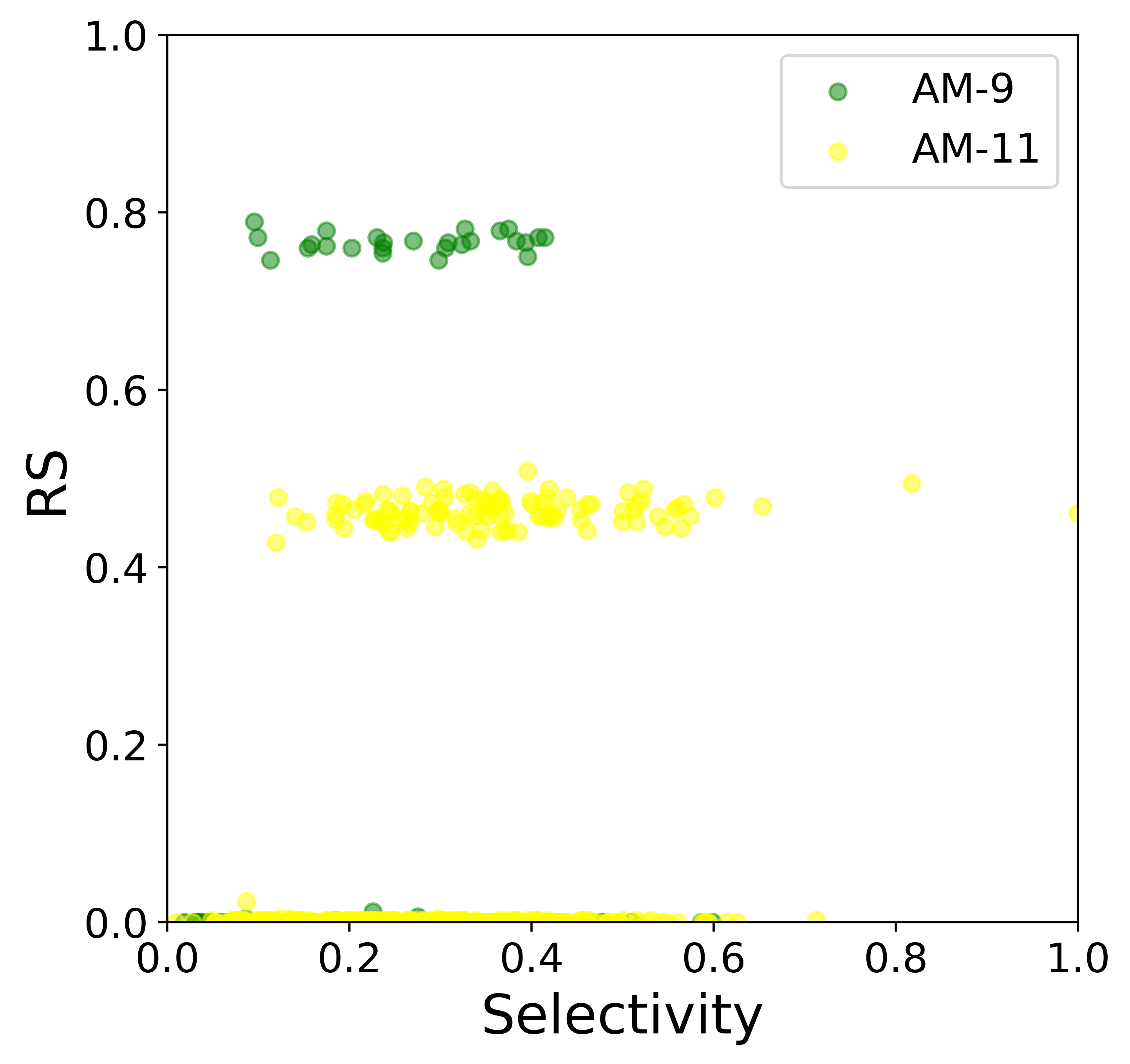}}
  \caption{\textbf{The convolutional layers exhibit the functions of feature extraction and integration.} (a) shows the RS and class selectivity of the first and third layers. (b) is for the 9th and 11th layers.}
  \label{fig:vggcifar} 
\end{figure}

\begin{figure}[tbp]
  \centering
  \subfigure[layer1]{
    \label{fig:subfig:a} 
    \includegraphics[scale=0.22]{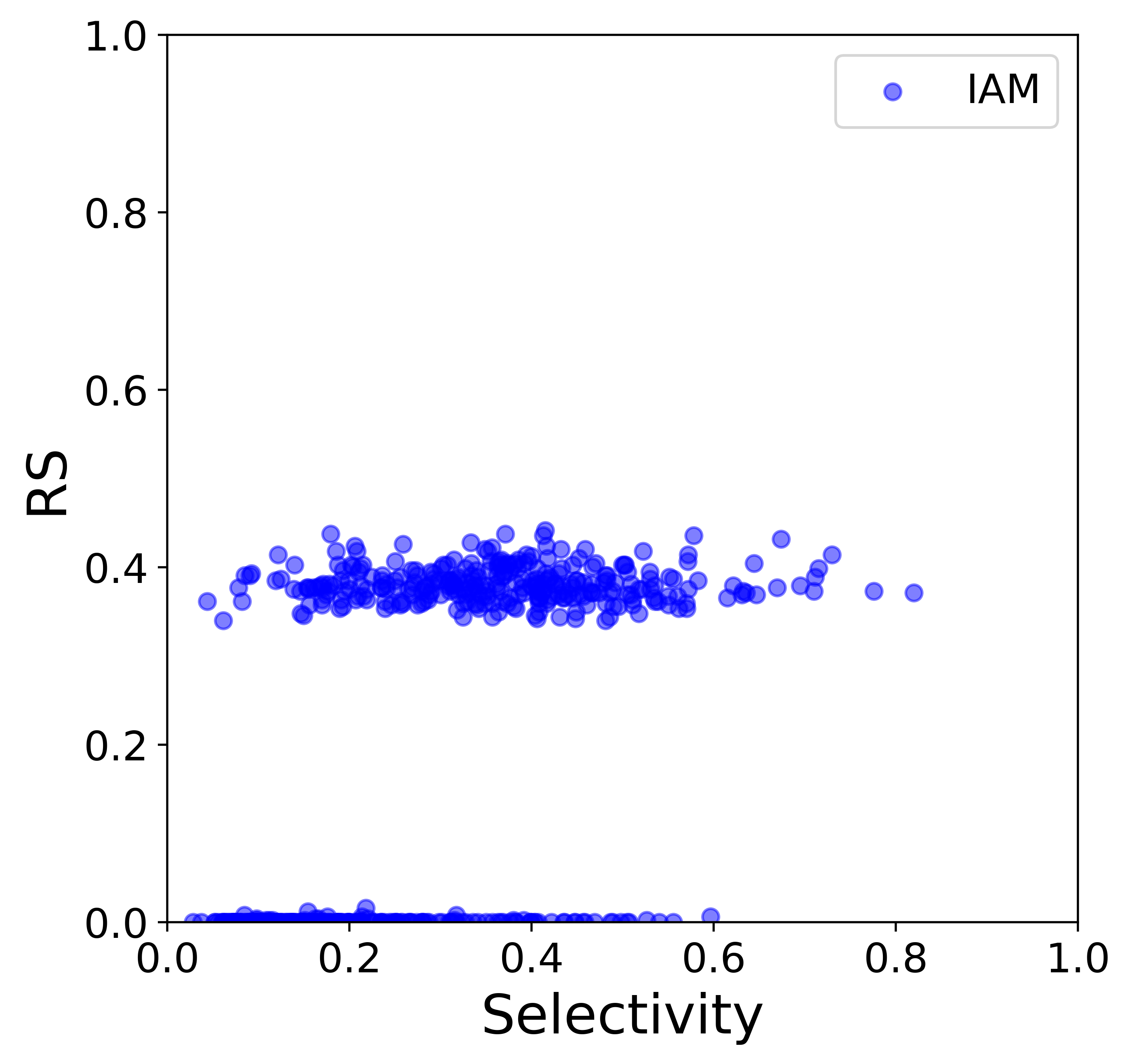}}
  \subfigure[layer3]{
    \label{fig:subfig:b} 
    \includegraphics[scale=0.22]{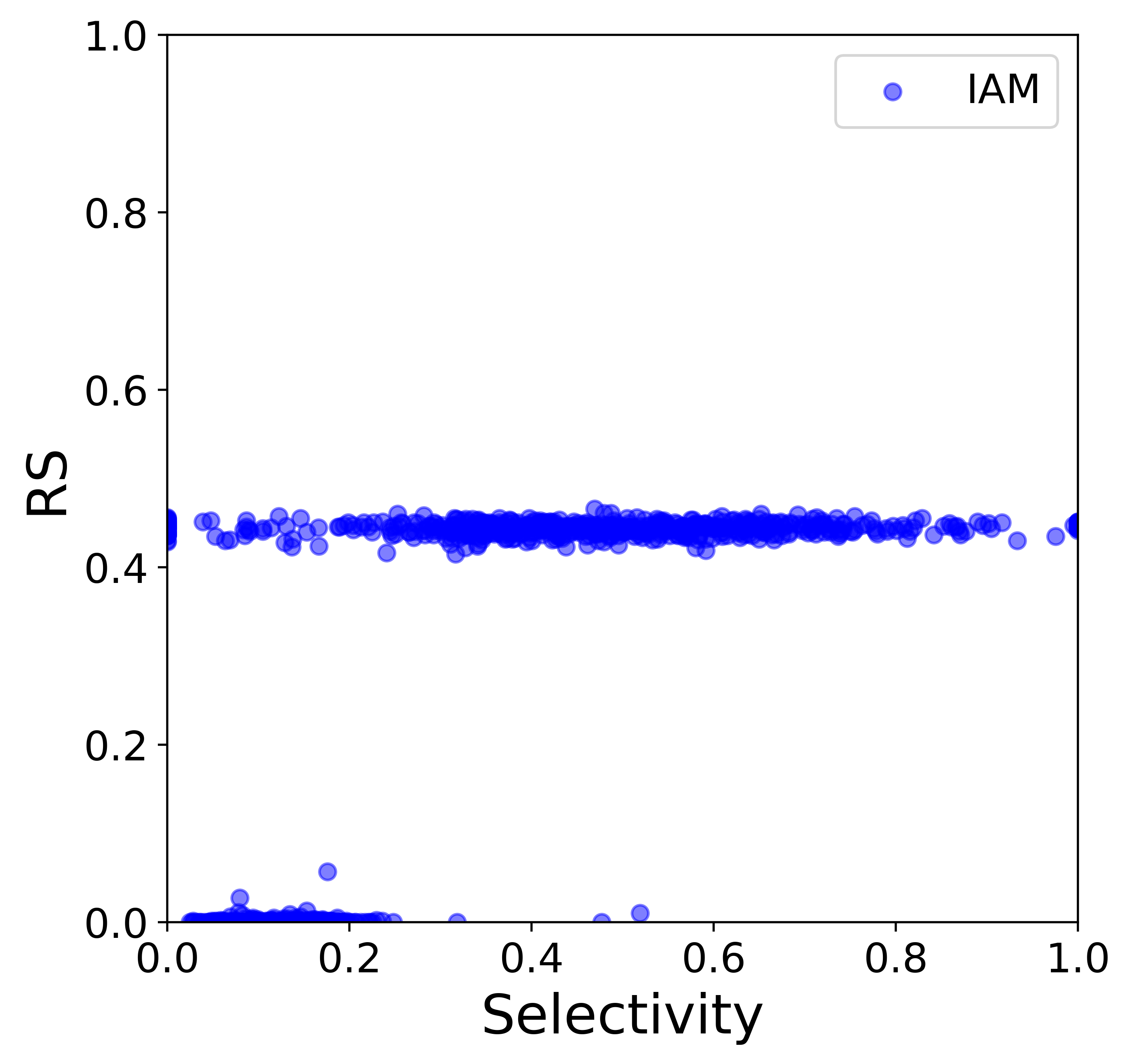}}
    \caption{\textbf{The MLP does not exist in the diversity of the units' RS.} Relation between the RS and class selectivity in the first layer of the MLP (a) and the third layer (b).}
  \label{fig:mlp} 
\end{figure}

We conduct the same experiments on VGG16 trained on Cifar-10, as shown in Fig.\ref{fig:vggcifar}. Similarly, we conclude that WIUs have low RS in shallow layers. However, the units’ RS presents a small value and no longer has diversity in deep layers. This phenomenon starts from a certain layer, and its division of the network layer is evident. The first few layers of the network are mainly used for feature extraction, and the main functions of the latter layers are feature integration for classification.

We conduct experiments on an MLP trained on Cifar-10, as shown in Fig.\ref{fig:mlp}. The RS is mostly the same and does not change considerably in all layers. Depending on previous conclusions, the function of the MLP is feature integration. The MLP is prone to overfitting and poor generalization. This finding explains that MLP performs poorly because it has no layer to extract features.

\begin{figure}[tbp]
  \centering
  \subfigure[layer3 and layer11]{
    \label{fig:vgg3-11} 
    \includegraphics[scale=0.22]{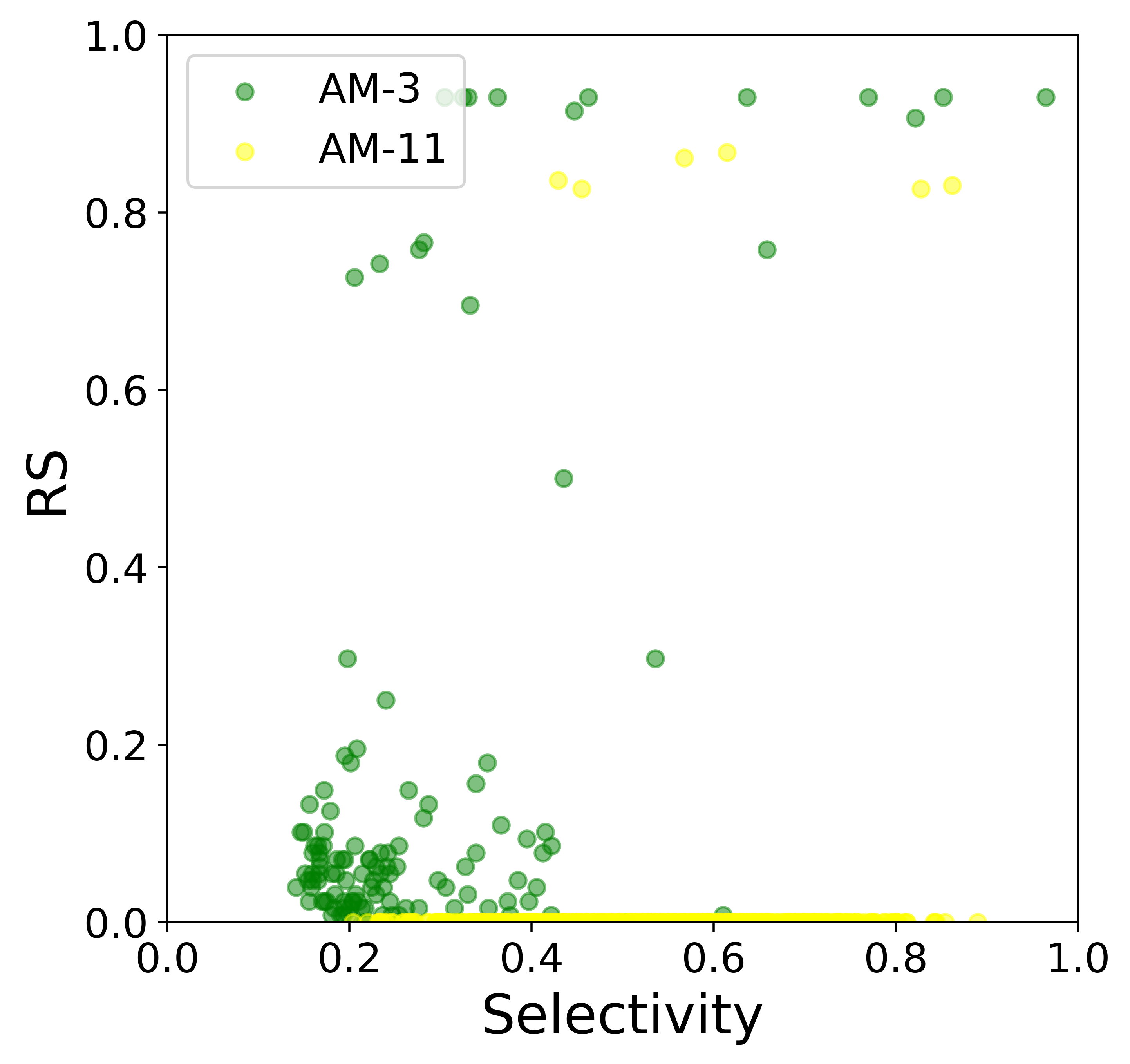}}
  \subfigure[correlation between the RS and class selectivity under different layers]{
    \label{fig:vgg-r} 
    \includegraphics[scale=0.22]{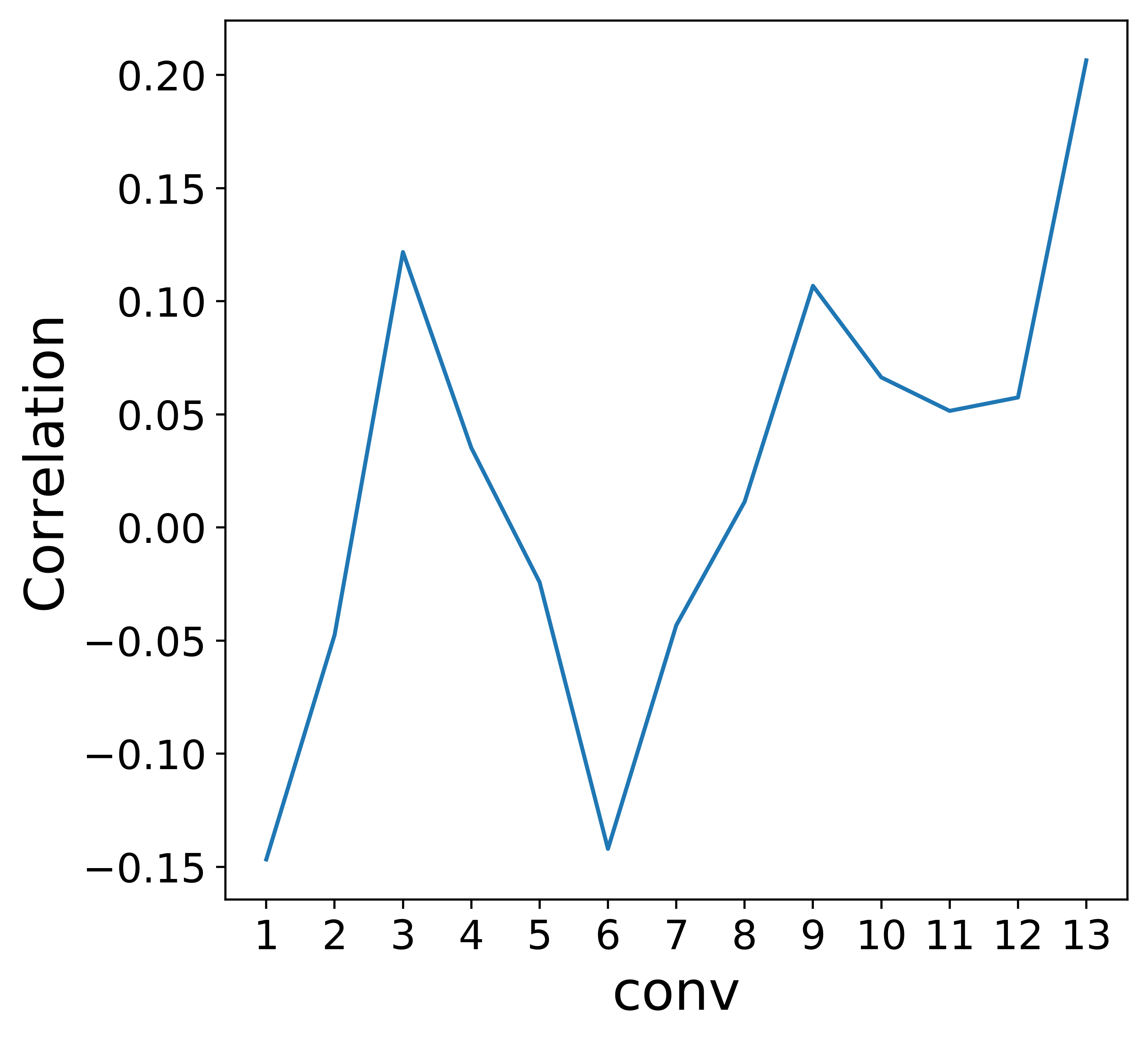}}
    \caption{\textbf{The RS and class selectivity are related in part layers under the ImageNet dataset.} (a) Relation between the RS and class selectivity in the 3rd and 11th layers (b) The correlation of the RS and class selectivity is irregular in all convolutional layers.}
  \label{fig:vgg16} 
\end{figure}

We use the pre-trained VGG16 model and ImageNet to perform the same experiments, as shown in Fig.\ref{fig:vgg3-11}. The results imply that a large number of WIUs assemble at the bottom left of the figure, which provide many independent representations of the network. The RS and class selectivity in all layers are shown in Fig.\ref{fig:vgg-r}. The RS and class selectivity show no correlation. This finding differs from our observation. Further work will need to consider the impacts of RS and class selectivity on the network.

\section{Conclusions}
This study clarifies the insignificant role of HIUs in DNNs from a feature space perspective, which inspires us to pay considerable attention to the independence and relationship of representations for explaining the working mechanism of DNNs. Convolutional layers actually have distinct functions, although they have the same structure. We need to consider the impact of the network structure when we study the interpretability of the network. However, the correlation between the RS and HIUs in a large dataset is not evident at the end of the experiment. Using a single metric to understand the internal representations of DNNs is thus unreasonable. Considering that negative values of images are meaningless, we analyze outputs by positive values. Nevertheless, negative activation in neuroscience can indicate an inhibition, which may be considered in the future.

\bibliography{references} 
\bibliographystyle{aaai}

\end{document}